\documentclass[sigconf]{aamas}

\usepackage[linesnumbered,ruled,vlined]{algorithm2e}
\usepackage{bbm}

\newcommand{\E}{\mathbb{E}}

\newcommand{\cl}{c_{\mathrm{lane}}}
\newcommand{\cc}{c_{\mathrm{coll}}}

\newcommand{\drcpo}{CRLLK-D}
\newcommand{\crcpo}{CRLLK-C}

\newcommand{\dppof}{PPO-D}
\newcommand{\cppof}{PPO-C}

\usepackage{balance} 
\usepackage{boldline}
\usepackage{makecell}
\usepackage{multirow} 
\usepackage{subcaption}

\makeatletter
\gdef\@copyrightpermission{
  \begin{minipage}{0.2\columnwidth}
   \href{https://creativecommons.org/licenses/by/4.0/}{\includegraphics[width=0.90\textwidth]{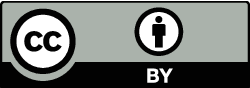}}
  \end{minipage}\hfill
  \begin{minipage}{0.8\columnwidth}
   \href{https://creativecommons.org/licenses/by/4.0/}{This work is licensed under a Creative Commons Attribution International 4.0 License.}
  \end{minipage}
  \vspace{5pt}
}
\makeatother

\setcopyright{ifaamas}
\acmConference[AAMAS '25]{Proc.\@ of the 24th International Conference
on Autonomous Agents and Multiagent Systems (AAMAS 2025)}{May 19 -- 23, 2025}
{Detroit, Michigan, USA}{Y.~Vorobeychik, S.~Das, A.~Nowé  (eds.)}
\copyrightyear{2025}
\acmYear{2025}
\acmDOI{}
\acmPrice{}
\acmISBN{}

\acmSubmissionID{23}

\title[AAMAS-2025 Formatting Instructions]{CRLLK: Constrained Reinforcement Learning for Lane Keeping in Autonomous Driving}

\subtitle{Demonstration Track}

\author{Xinwei Gao}
\affiliation{
  \institution{Nanyang Technological University}
  \country{Singapore}}
\email{xinwei.gao@ntu.edu.sg}

\author{Arambam James Singh}
\affiliation{
  \institution{Indian Institute of Technological}
  \country{Delhi, India}}
\email{jamesa@iitd.ac.in}

\author{Gangadhar Royyuru}
\affiliation{
  \institution{Indian Institute of Technology}
  \city{Madras}
  \country{India}}
\email{rspgangadharroyyuru@gmail.com}

\author{Michael Yuhas}
\affiliation{
  \institution{Nanyang Technological University}
  \country{Singapore}}
\email{michaelj004@e.ntu.edu.sg}

\author{Arvind Easwaran}
\affiliation{
  \institution{Nanyang Technological University}
  \country{Singapore}}
\email{arvinde@ntu.edu.sg}

\begin{abstract}
Lane keeping in autonomous driving systems requires scenario-specific weight tuning for different objectives. We formulate lane-keeping as a constrained reinforcement learning problem, where weight coefficients are automatically learned along with the policy, eliminating the need for scenario-specific tuning. Empirically, our approach outperforms traditional RL in efficiency and reliability. Additionally, real-world demonstrations validate its practical value for real-world autonomous driving.

\end{abstract}


\newcommand{\BibTeX}{\rm B\kern-.05em{\sc i\kern-.025em b}\kern-.08em\TeX}

\begin{document}


\pagestyle{fancy}
\fancyhead{}


\maketitle 


\section{Introduction}
\begin{figure*}[ht]
	\centering
	\begin{subfigure}[b]{0.23\textwidth}
	\centering
		\includegraphics[width=3.9cm, height=2.5cm]{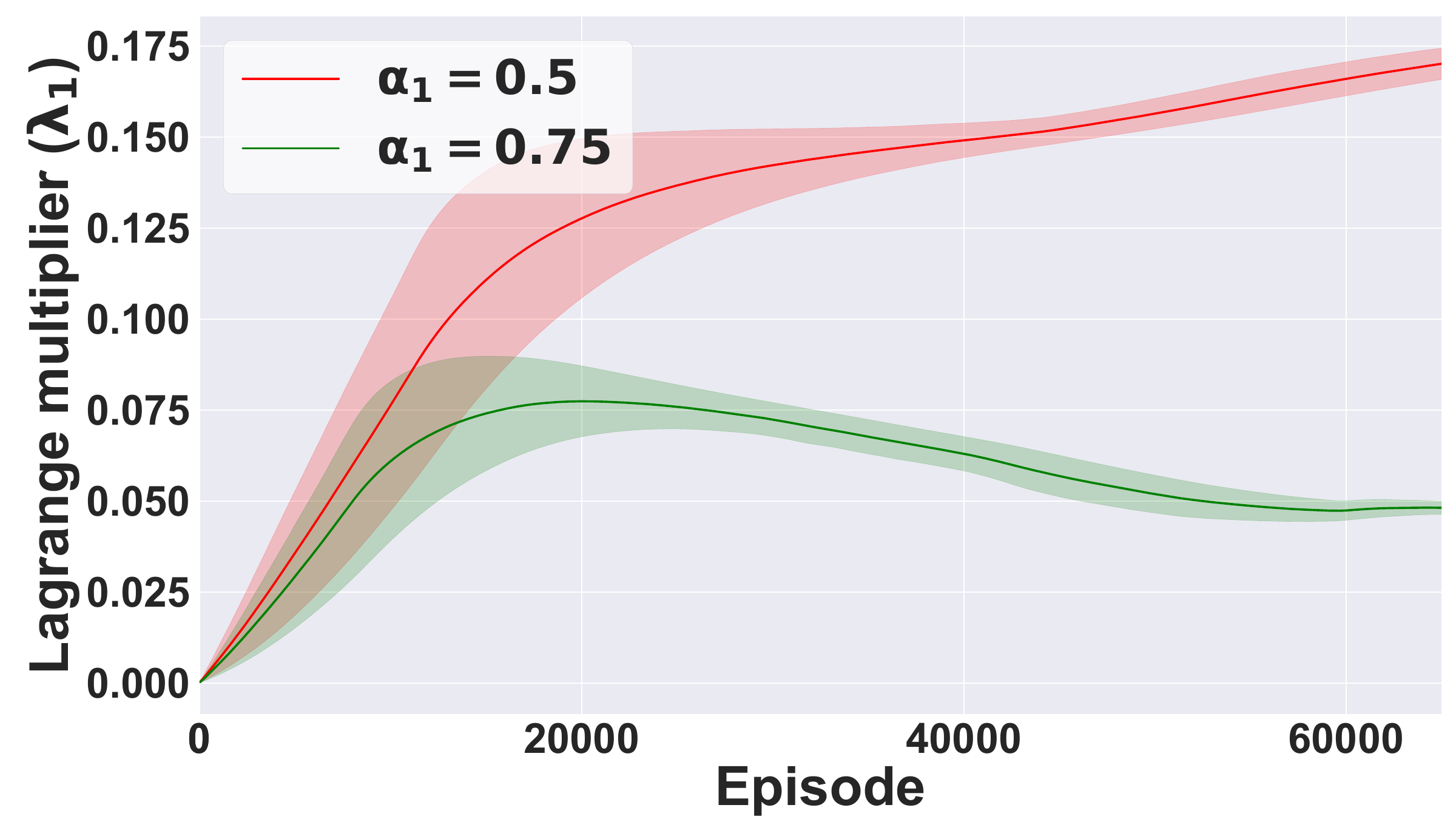}
        \vspace{-1.5mm}
		\caption{Langrangian multiplier $\lambda_{1}$}
		\label{fig:lc_lambda}
	\end{subfigure}
	\hspace{0.005\textwidth}
    \begin{subfigure}[b]{0.23\textwidth}
	    \centering
		\includegraphics[width=3.9cm, height=2.5cm]{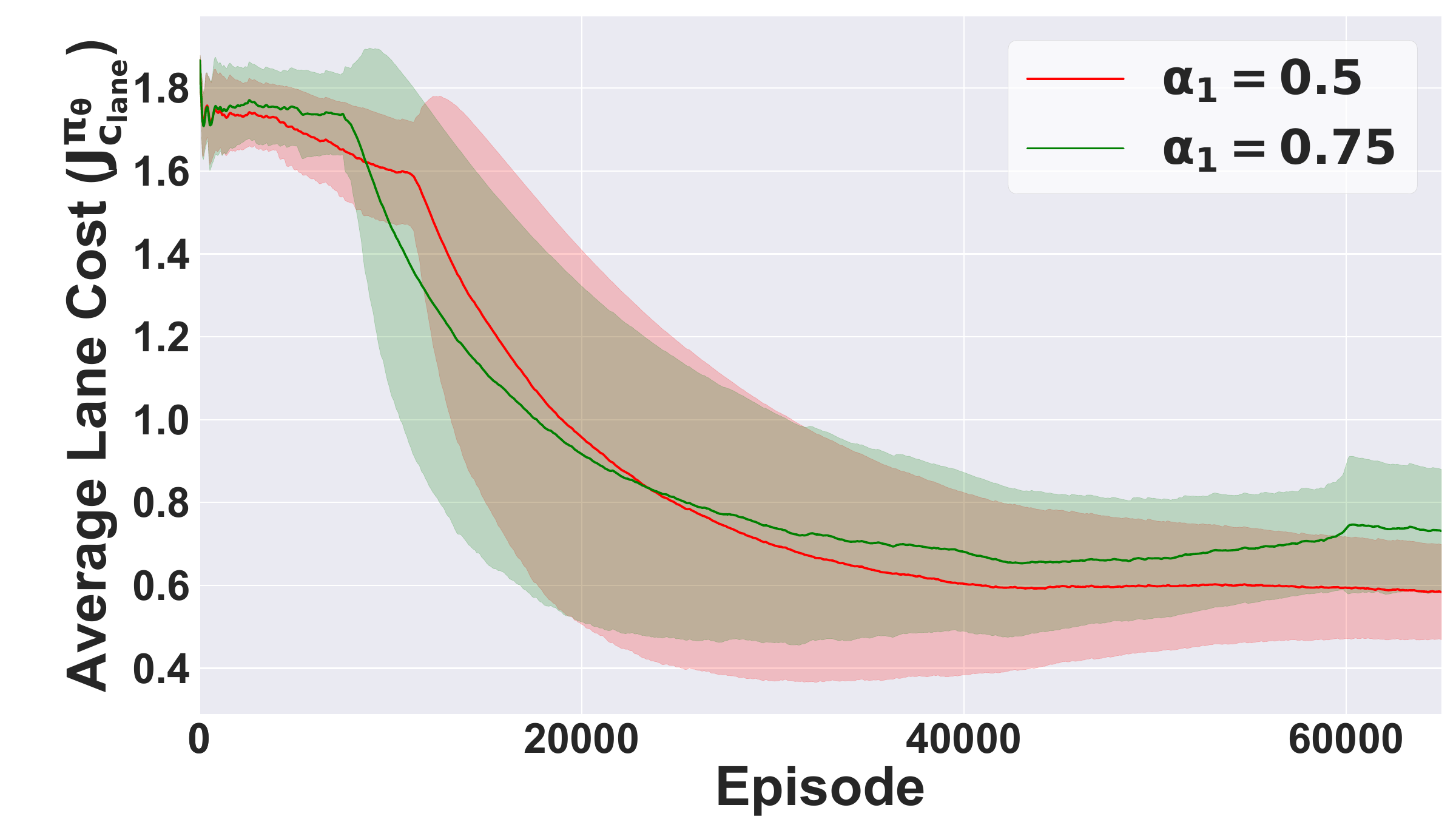}
        \vspace{-1.5mm}
		\caption{Average lane deviation $J_{\cl}^{\pi_\theta}$}
		\label{fig:lc_jclane}
	\end{subfigure}%
	\begin{subfigure}[b]{0.23\textwidth}
	\centering
		\includegraphics[width=3.9cm, height=2.5cm]{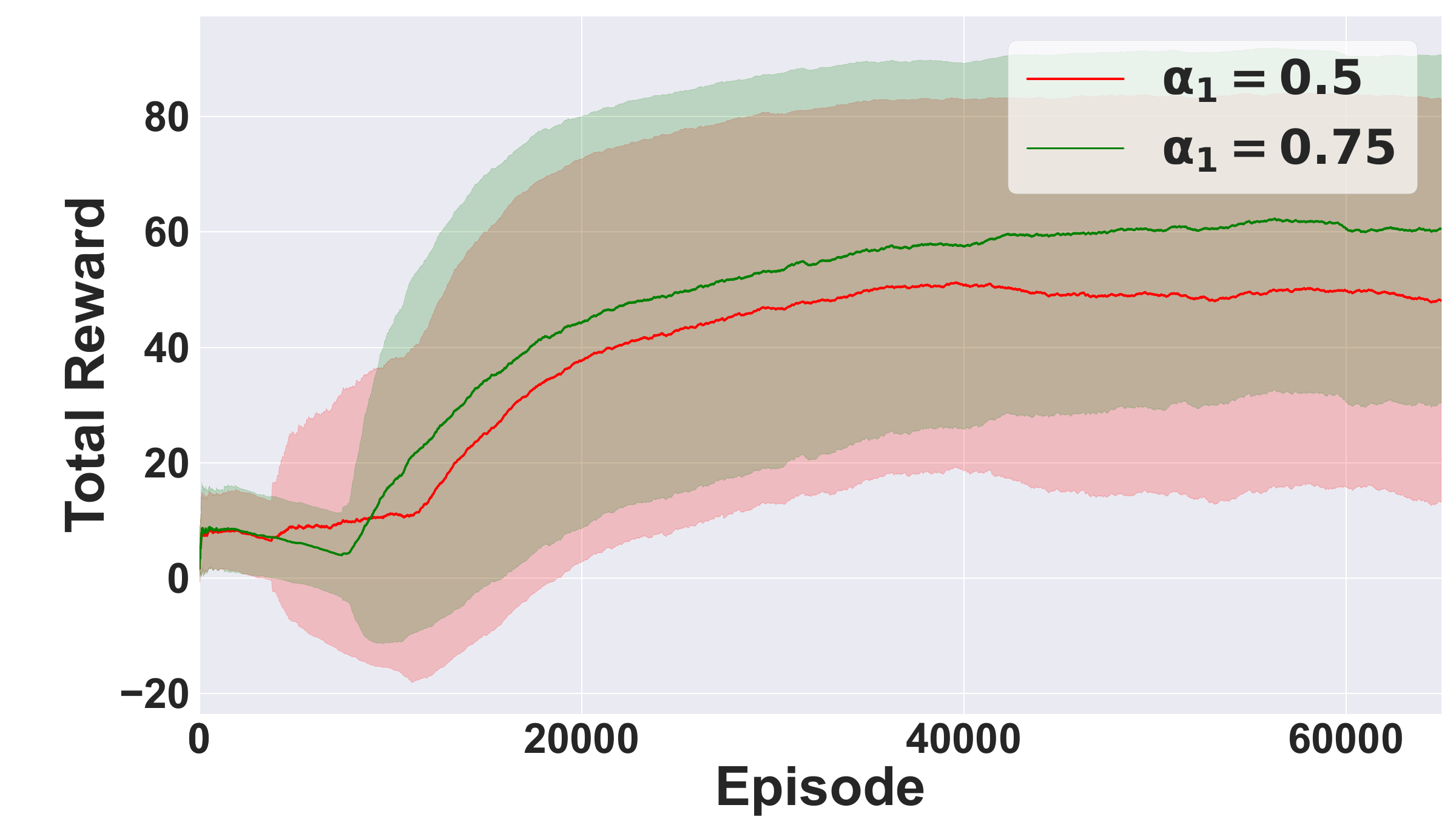}
        \vspace{-1.5mm}
		\caption{Total distance reward $J_R^{\pi_\theta}$}
		\label{fig:lc_return}
	\end{subfigure}
    \begin{subfigure}[b]{0.29\textwidth}
	    \centering
		\includegraphics[width=5.0cm, height=2.5cm]{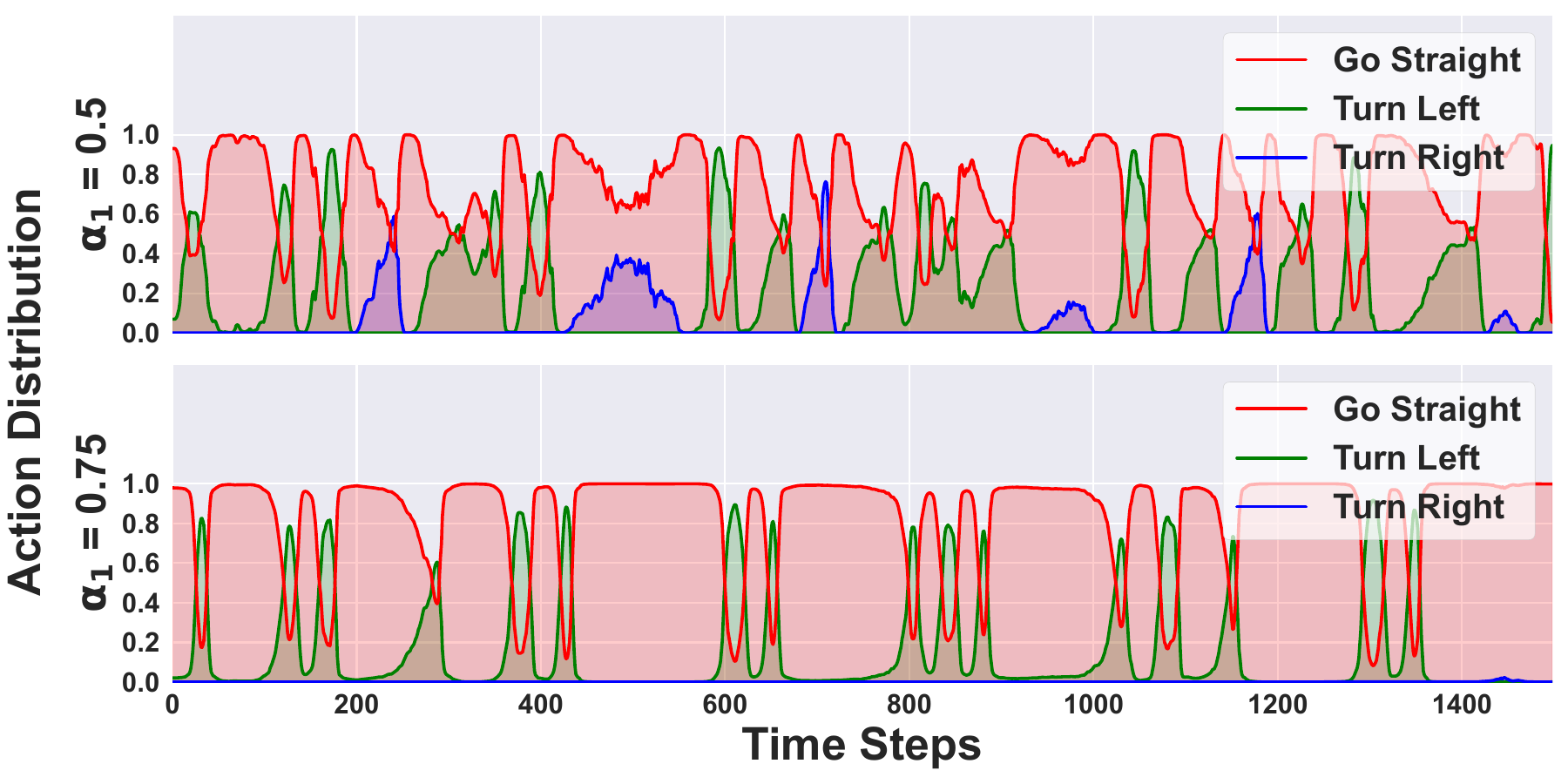}
        \vspace{-1.5mm}
		\caption{Action distribution in evaluation.}
		\label{fig:ab_actionlog}
	\end{subfigure}
    \vspace{-7mm}
    \caption{(a-c) Learning curves for different constraint threshold values ($\alpha_1$). (d) Action distribution for one test episode.}
    \vspace{-4mm}
    \label{fig:learningandactions}
\end{figure*}

The problem of lane keeping (LK) is an instance of a challenging real-time sequential decision-making problem in the domain of self-driving cars or autonomous driving systems~\cite{chen2020lane,netto2004lateral,sallab2016end}. 
Traditional model-free reinforcement learning (RL) based approaches to the LK problem face the challenge of defining a reward function that manages trade-offs between multiple objectives~\cite{claussmann2019review}. 
Prior RL approaches use fixed-weighted combinations for objectives such as driving distance~\cite{almasi2020robust}, minimizing yaw angle~\cite{feher2018q, kHovari2020design}, crash avoidance~\cite{ashwin2023deep, yuan2020race}, and lateral/longitudinal control~\cite{duan2020hierarchical, muhammad2020deep}.
Traditional multi-objective reinforcement learning (MORL) approaches~\cite{van2014multi, hayes2022practical} tackle such problems by learning a set of optimal policies or applying scalarized reward schemes that rely on static weighting. However, a fundamental limitation of such approaches is that the weight coefficients are typically obtained using scenario-specific tuning and extensive grid searches, which is both time-consuming and computationally expensive for high-fidelity physical simulators. 

In response to this challenge, this work introduces a constrained reinforcement learning based formulation and learning approach for the LK problem, which dynamically adjusts the weight coefficients of different objectives. By leveraging this constrained formulation, our system significantly outperforms traditional approaches in terms of efficiency (travel distance) and reliability (lane deviations and avoidance of collisions). We present the weight coefficient learning process and validate our framework in both simulation and real-world settings.\footnote{Demonstration video: \url{youtu.be/1BlwJOIUaGM}}\footnote{Source code: \url{github.com/CPS-research-group/CPS-NTU-Public/tree/AAMAS2025}}
\section{Constrained RL for Lane Keeping}



    
    



\subsection{Problem Formulation}

Lane keeping is considered as a multi-objective problem with reward and cost functions defined at each time step. The \textit{travel distance reward} is $r(s_t, a_t) = \mathrm{d}^\mathrm{trv}$, where $\mathrm{d}^\mathrm{trv}$ denotes the forward distance. The \textit{lane deviation cost}, $\cl(s_t, a_t) = \mathrm{d}^\mathrm{lane}$, penalizes horizontal deviation from the lane center, with the average performance given by $J_{\cl}^{\pi_\theta} = \E \big[\frac{1}{H}\sum_{t=0}^{H-1} \cl(s_t, a_t)\big]$. The \textit{collision cost}, $\cc(s_t, a_t) = \mathbf{1}^{\mathrm{col}}$, takes the value $1$ if the agent collides with obstacles or boundaries, with performance $J_{\cc}^{\pi_\theta} = \E \big[\sum_{t=0}^{H-1} \cc(s_t, a_t)\big]$ across multiple episodes.
To ensure safe and efficient driving, we formulate lane-keeping as a constrained optimization problem:
\begin{align}
\max _{\pi_\theta \in \Pi} J_R^{\pi_\theta},\;\; 
\text {s.t. } \;\; J_{\cl}^{\pi_\theta} \leq \alpha_{1},\; J_{\cc}^{\pi_\theta} \leq \alpha_{2},
\label{eq:cmdp2}
\end{align}
where $\alpha_{1}$ and $\alpha_{2}$ are non-negative thresholds for costs. $\alpha_1$ corresponds to a real-world distance in decimeters ($10^{-1}m$).

We apply Lagrangian relaxation (LR) technique~\cite{bertesekas1999nonlinear} to convert the constrained optimization in Equation~\eqref{eq:cmdp2} into an equivalent unconstrained optimization problem as follows:
\begin{align}
\min_{\lambda_i \geq 0} \max_\theta L(\lambda_1, \lambda_2, \theta) = \min_{\lambda_i \geq 0} \max_\theta \Big[J_R^{\pi_\theta} - \sum_{i=1}^{2} \lambda_i \left(J_{C_i}^{\pi_\theta} - \alpha_i\right)\Big],
\label{eq:cmdpd}
\end{align}
where $i \in \{1, 2\}$, $L$ is Lagrangian, $\lambda_i$ are Lagrangian multipliers for lane and collision costs. $\lambda_i$ and $\theta$ are updated following the gradient descent and ascent approach as follows:
\begin{alignat}{1}
\lambda_i^{\mathrm{new}} &= \max\left(0, \lambda_i^{\mathrm{old}} - \eta_i \left(J_{C_i}^{\pi_\theta} - \alpha_i\right)\right), \quad i \in \{1, 2\},\label{eq:grlambdai}\\
\theta^{\mathrm{new}} &=\theta^{\mathrm{old}}+\eta_{3} \nabla_\theta \E \Big[\log \pi_\theta(a|s)\Big(J_R^{\pi_\theta} - \lambda_{1} J_{C_{\mathrm{lane}}}^{\pi_\theta} - \lambda_{2} J_{C_{\mathrm{coll}}}^{\pi_\theta}\Big)\Big], \label{eq:theta_combined}
\end{alignat}
where $\eta_{1}$, $\eta_{2}$ and $\eta_{3}$ are learning rates. Equations~\eqref{eq:grlambdai} updates $\lambda_{i}$ to satisfy the original cost constraints. The policy parameters $\theta$ is updated with the discounted cost version of $J_{c_{\mathrm{lane}}}^{\pi_{\theta}}$ and $J_{c_{\mathrm{coll}}}^{\pi_{\theta}}$ to track the recursive property of Bellman equation~\cite{bellman1952theory}, where $J_{c_{\mathrm{lane}\cdot\gamma}}^{\pi_{\theta}} \triangleq\mathbb{E}\left[\sum_{t=0}^{H-1} \gamma^t \cl(s_t, a_t) \right]$ and $J_{c_{\mathrm{coll}\cdot\gamma}}^{\pi_{\theta}} \triangleq\mathbb{E}\left[\sum_{t=0}^{H-1} \gamma^t \cc(s_t, a_t) \right]$.
Using the discounted cost constraints above, we define a modified reward function that applied in our approach:
\begin{alignat}{1}
\hspace{-0.12cm}\hat{r}( s_t, a_t, \lambda_{1}, \lambda_{2}) 
 \triangleq  r(s_t,a_t) - \lambda_{1} \cl(s_t,a_t) - \lambda_{2} \cc(s_t,a_t) \label{eq:rcpo_reward}.
\end{alignat}
In the reward function $\hat{r}(\cdot)$, parameters $\lambda_1$ and $\lambda_2$ act as the adaptive weight coefficients of objectives.

\subsection{Implementation}
Existing constraint RL solutions like RCPO~\cite{TesslerMM19} update policy with truncated samples, failing to track collision cost constraints across multiple episodes, while two-timescale frameworks are impracticable for high-fidelity physical simulators due to computational costs. 
We adopt a one-timescale framework, updating both $\theta$ and $(\lambda_{1},\lambda_{2})$ simultaneously, akin to the simplifications of Actor-Critic~\cite{konda1999actor} made by A3C~\cite{mnih2016asynchronous} and DDPG~\cite{silver2014deterministic}. We implement this approach on the Duckietown platform with PPO~\cite{schulman2017proximal} algorithm. The system observes the vehicle's state through camera and computes actions while minimizing lane deviation and avoiding obstacles.

\section{Experiment and Demonstration}
\subsection{Simulation Evaluation}
The simulation evaluation results on a \textit{small loop} scenario are listed in Table~\ref{tab:tab_sl_zz}. \drcpo, \crcpo\ are implementations of our approach on discrete and continuous setting, while \dppof, \cppof\ are the traditional approach with a fixed reward in~\cite{schulman2017proximal}. Each method is evaluated on 100 test episodes and averaged over two different seeds. 
Further evaluation results for baselines and scenarios are listed in the video. This evaluation result shows significant performance improvement by adapting our constrained RL framework in terms of efficiency (higher $J_{R}$ for travel distance), and reliability (lower $J_{\cl}$ for lane deviations and lower $J_{\cc}$ for collision). 

\begin{table}[h]
    \centering
    \vspace{-0mm}
    \caption{Performance comparison on the small loop scenario with $\alpha_1 = 0.5$ and $\alpha_2 = 0.02$.}
    \small 
    \vspace{-2.5mm}
    \begin{tabular}{ccccccccc}
\hlineB{2}
\addlinespace[2pt]
 & \drcpo & \dppof & \crcpo & \cppof \\
\midrule
$J_{\cl}$ & 0.66$\pm$0.02 & 0.99$\pm$0.07 & \textbf{0.31$\pm$0.00} & 1.04$\pm$0.25 \\
$J_{\cc}$ & 0.17$\pm$0.00 & 0.38$\pm$0.04 & \textbf{0.05$\pm$0.01} & 0.34$\pm$0.20 \\
$J_R$ & \textbf{69.0$\pm$0.6} & 46.6$\pm$3.2 & 62.4$\pm$22.2 & 51.3$\pm$14.8 \\
\addlinespace[2pt]
\hlineB{2}
\end{tabular}
    \label{tab:tab_sl_zz}
    \vspace{-1.5mm}
\end{table}
\begin{figure}[h]
	\centering
    \hspace{0.00\textwidth}
	\begin{subfigure}[b]{0.15\textwidth}
	    \centering
		\includegraphics[width=2.2cm, height=2.0cm]{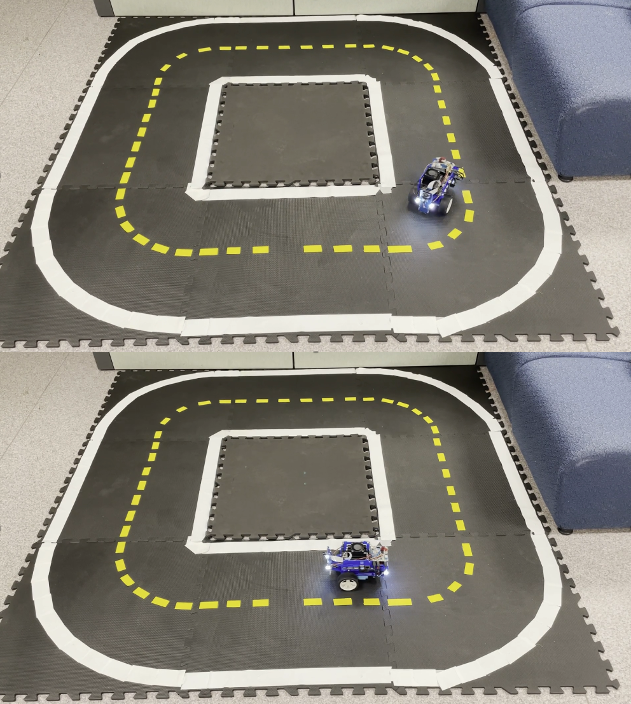}
        \vspace{-1mm}
		\caption{Small loop}
	\end{subfigure}%
    \hspace{0.00\textwidth}
	\begin{subfigure}[b]{0.15\textwidth}
	    \centering
		\includegraphics[width=2.2cm, height=2.0cm]{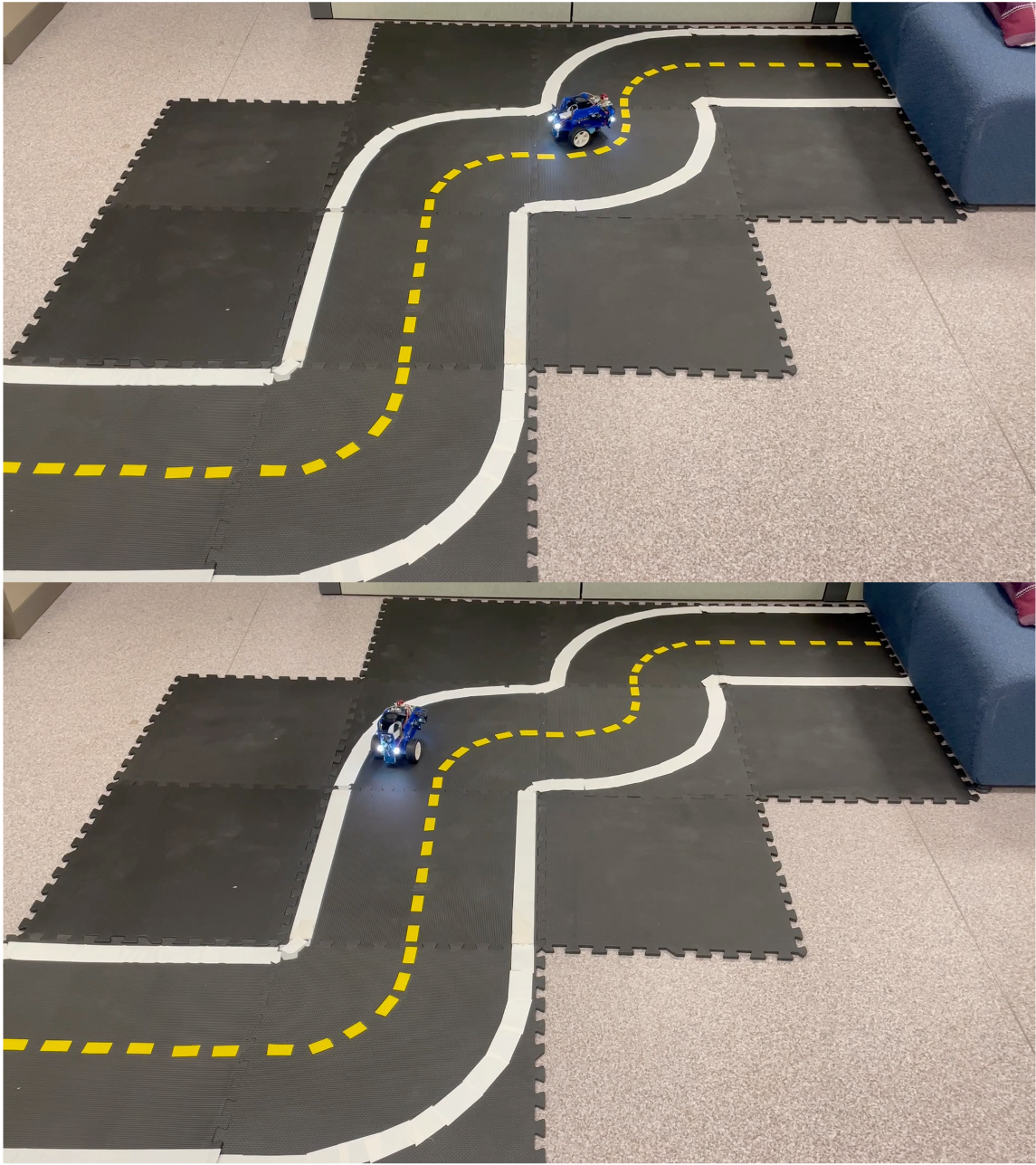}
        \vspace{-1mm}
		\caption{Zig-zag}
	\end{subfigure}%
	\hspace{0.00\textwidth}
	\begin{subfigure}[b]{0.15\textwidth}
	\centering
		\includegraphics[width=2.2cm, height=2.0cm]{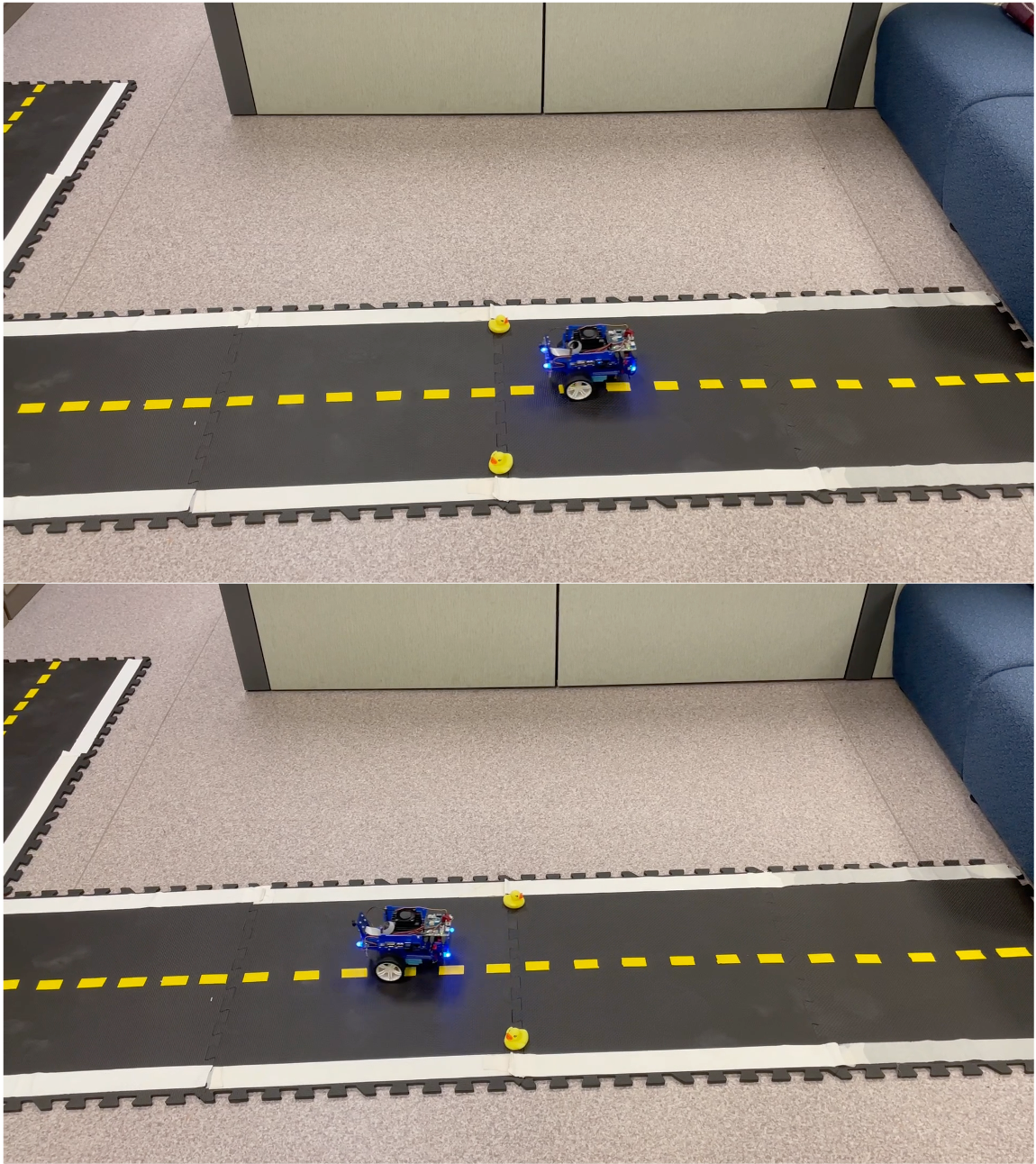}
        \vspace{-1mm}
		\caption{Obstacle loop}
	\end{subfigure}
    \vspace{-4mm}
	\caption{Real-world evaluation in scenarios.}
	\label{fig:real}
    \vspace{-5.0mm}
\end{figure}

We performed a demonstration on the training convergence and testing behavior for \drcpo\ on a \textit{small loop} map in Figure~\ref{fig:learningandactions}, with two different constraint levels: tight constraint ($\alpha_1=0.5$) and loosened constraint ($\alpha_1=0.75$). For tight constraint, the agent policy has a higher probability to choose turn-left and turn-right actions than in the loose constraint, showing fewer lane deviations but covering a shorter total distance. Through this study, we gain insight into how weight coefficients are dynamically learned and how different constraint thresholds affect the lane-keeping trade-off between the lane deviation and travel distance.


\subsection{Real-World Demonstration}
We evaluate the constrained RL-based policy for lane-keeping tasks using a Duckiebot in real-world scenarios after training in simulation. Testing covered three scenarios: low-difficulty (\textit{small loop}), high-difficulty (\textit{zig-zag}), and complex (\textit{obstacle loop}). 
The Duckiebot, equipped with Jetson Nano~\cite{suzen2020benchmark} hardware and ROS2~\cite{macenski2022robot}, uses an autonomous driving architecture~\cite{velasco2020autonomous} for perception, decision-making, and control. Despite differences between simulation and real-world conditions (e.g. lighting, obstacles), the Duckiebot adapts and navigates smoothly. Figures~\ref{fig:real} showcase snapshots of the experiments, with results detailed in the demonstration video.

\section{Conclusion}
\label{sec:concl}
We formulate lane keeping as a constrained optimization problem and propose a constrained RL-based solution. The weight coefficients are adaptively learned, eliminating the need for scenario-specific tuning. Empirically, our approach outperforms traditional RL. We analyze the impact of constraint thresholds on policy behavior and convergence, while validating our method through real-world demonstrations across various scenarios.
\begin{acks}
This research is supported by the National Research Foundation, Singapore and DSO National Laboratories under the Al Singapore Programme (AISG Award No: AISG2-RP-2020-017), and MoE, Singapore, Tier-2 grant number MOE2019-T2-2-040.
\end{acks}


\clearpage
\bibliographystyle{ACM-Reference-Format} 
\bibliography{main}


\end{document}